\documentclass[runningheads]{llncs}
\usepackage[utf8]{inputenc}
\usepackage[T1]{fontenc}
\usepackage{amsmath,amssymb}
\usepackage{graphicx}
\usepackage{booktabs}
\usepackage{cite} 
\usepackage{hyperref}
\usepackage{algorithm}
\usepackage{algpseudocode}
\usepackage{multirow}
\usepackage{xcolor}

\begin{document}

\title{A Deep Reinforcement Learning Framework for Real-Time Electric Vehicle
Charging Management in IoEV-Enabled Smart Cities}

\author{Nour-Eddine Moumni\inst{1}\orcidID{0009-0008-5803-0604} \and
        Rachid Alaoui\inst{1,3}\orcidID{0000-0001-9948-329X} \and
        Driss Kiouach\inst{2}\orcidID{0000-0002-1497-3028}}

\authorrunning{Moumni et al.}

\institute{
  Laboratory of Research in Computer Science and Telecommunications (LRIT),\\
  Mohammed V University, Rabat, Morocco\\
  \email{noureddine\_moumni@um5.ac.ma}
  \and
  Laboratory of Probability, Statistics and Intelligent Systems (LPAIS),\\
  Sidi Mohamed Ben Abdellah University, Fez, Morocco
  \and
  Higher School of Technology of Sal\'e (EST Sal\'e),\\
  Mohammed V University, Rabat, Morocco
}

\maketitle

\begingroup
\renewcommand\thefootnote{}
\footnotetext{This is the authors' accepted manuscript (paper ID
733916), accepted as a regular paper and presented at the 10th
International Conference on Advanced Intelligent Systems for
Sustainable Development (AI2SD'2025 cycle), carried over for
publication in the AI2SD 2026 proceedings, scheduled to appear in an
indexed Springer \emph{Lecture Notes in Networks and Systems} volume.
This version is posted as a preprint ahead of that publication.}
\endgroup

\begin{abstract}
The rapid growth of electric vehicles (EVs) in urban environments demands
intelligent, decentralised charging management strategies that operate under
strict real-time constraints. This paper proposes an \emph{Edge-Intelligent}
architecture in which each EV would select a charging station (CS) by
querying a shared, centrally trained Deep Q-Network (DQN) policy over a
lightweight MQTT publish--subscribe backbone; independent per-EV edge
inference is the deployment target this architecture is designed for, not
the configuration evaluated here. The reported simulation evaluates that
shared DQN policy applied sequentially to each vehicle's observations
within a single SUMO run, using MQTT to carry KPI telemetry while
observations are obtained through TraCI. The system is evaluated on two real-world
road networks extracted from OpenStreetMap (Rabat Agdal--Hay Riad and Tangier,
Morocco); the primary network carries four geolocated public charging stations
offering heterogeneous power levels (50\,kW and 150\,kW). Three traffic demand profiles (morning, midday, and
evening peak) are simulated using SUMO, and four decision policies are
compared: an Adaptive Heuristic baseline, a Tabular Q-Learning agent, the
proposed DQN agent, and a Double DQN variant. A multi-seed statistical protocol (7 seeds,
Wilcoxon signed-rank test, Cohen's~$d$) evaluates the results. The DQN
agent achieves a 6.8\% relative improvement in charge rate and a 5.4\% increase
in delivered energy over the Adaptive Heuristic under Evening Peak demand
($p = 0.016$, Cohen's $d = 0.99$ and $1.51$), while showing no advantage in
waiting time; on the second network the heuristic in turn leads on waiting
time, so the ranking is not stable across geographies. A dedicated scalability analysis shows linear
growth of the state space with the number of stations and sub-millisecond
inference, and all runs are obtained under intermittent station outages
driven by a Markov failure process.

\end{abstract}

\section{Introduction}
\label{sec:intro}

Morocco has placed transport decarbonisation among its policy
priorities~\cite{itf2021}, with battery-electric-vehicle adoption
barriers analysed in their own right~\cite{nasreddin2023}; more
broadly, rapid worldwide EV-fleet growth strains urban charging
infrastructure and grid
stability~\cite{bhatti2024Electricvehiclecharging,moumni2025SmartEVrouting,sarda2024reviewelectricvehicle}.
Charging infrastructure nonetheless remains thin:
a survey of the public charging points listed for the greater Rabat area,
carried out in June 2026 against Google Maps and Chargemap (see
Section~\ref{sec:setup}), returned fewer
than twenty, creating acute risk of station congestion during peak commuting
hours. Two solution families have emerged: \emph{centralised}
approaches, which compute
globally optimal assignments but can incur communication latency
incompatible
with sub-second IoT requirements as system size grows~\cite{wan2019} ---
related to the coordinated charging-scheduling optimisation addressed
by Usman~et~al.~\cite{usman2021CoordinatedChargingScheduling} --- and \emph{rule-based}
heuristics, which react quickly but encode a fixed, non-adaptive
priority (distance or occupancy) whose adequacy depends on topology and
demand~\cite{moumni2026bentham}.

We address this gap through an \emph{Edge-Intelligent} architecture:
proposed for deployment, each EV would query a DQN policy, from station
observations exposed by the
environment, over an MQTT publish--subscribe
backbone~\cite{al-fuqaha2015InternetThingsSurvey,belli2020IoTEnabledSmartSustainable}, requiring no
centralised
controller at inference time. The evaluation reported in this paper runs that
policy shared and sequential within a single SUMO process, not as
independent per-EV instances (Section~\ref{sec:system}). The
policy maps a 13-dimensional continuous state vector---encoding battery state
of charge (SoC), distances to stations, occupancy levels, and queue
lengths---directly to a discrete action (choose CS$_k$,
$k \in \{1,2,3,4\}$).

The main contributions of this work are:
\begin{enumerate}
  \item A DQN-based charging assignment framework, designed for
        decentralised edge deployment and evaluated here as a shared
        policy in simulation, with a formal
        MDP formulation incorporating a queue-aware state representation.
  \item The first RL evaluation on real OpenStreetMap road networks in
        Morocco --- Rabat and Tangier --- using four verified real-world EV
        charging stations.
  \item An empirical comparison of four policies --- Adaptive Heuristic,
        Tabular Q-Learning, DQN and DDQN --- across three demand scenarios on
        two road networks, with statistically validated results (Wilcoxon
        signed-rank, $n=7$ seeds). The three structurally distinct policies
        carry the comparison; DDQN is reported as a variant of DQN, and the
        two are contrasted as a family in Section~\ref{sec:tanger}.
  \item A scalability and robustness analysis addressing large-scale
        deployment.
\end{enumerate}

The remainder of the paper is structured as follows.
Section~\ref{sec:related} reviews related work.
Section~\ref{sec:system} describes the system architecture.
Section~\ref{sec:mdp} formalises the MDP.
Section~\ref{sec:algos} details the three algorithms.
Section~\ref{sec:setup} describes the experimental setup.
Section~\ref{sec:results} presents and discusses results.
Section~\ref{sec:scalability} analyses scalability and robustness.
Section~\ref{sec:conclusion} concludes.

\section{Related Work}
\label{sec:related}

\subsection{RL-Based EV Charging Management}

RL and Deep RL (DRL) have been extensively investigated for EV charging
management across system scales, but centralised or grid-side works
assume centralised state observation and do not incorporate real-time
IoT communication constraints. Wan~et~al.~\cite{wan2019} pioneered DQN
for EV routing on synthetic networks; Ma~et~al.~\cite{ma2023PROLIFICDeepReinforcement},
Han~et~al.~\cite{han2024DQNbasedapproach}, and, in the related but
distinct domain of traffic-signal control,
Pang~et~al.~\cite{pang2024Scalablereinforcementlearning}, extend deep
Q-learning to large-scale scheduling under communication delays.
From a grid-side perspective, Da~Silva~et~al.~\cite{dasilva2020},
Cui~et~al.~\cite{cui2023Multiobjectiveoptimalscheduling}, and
Li~et~al.~\cite{li2024ReinforcementLearningEV} formulate multi-agent or
multi-objective DRL for station scheduling and renewable integration
without modelling vehicle mobility, while Lee and
Choi~\cite{lee2021} and
Michailidis~et~al.~\cite{michailidis2025ReinforcementLearningElectric}
take the operator's perspective with federated DRL for dynamic pricing
and solar/storage management.

Vehicle-side travel decisions have been addressed directly in smart-city
settings: Lee~et~al.~\cite{leekb2020} jointly select a route and
station; Farhani~et~al.~\cite{farhani2023} formulate assignment from
SUMO traffic data; Zhang~et~al.~\cite{zhang2021} recommend stations via
city-scale MARL; Chifu~et~al.~\cite{chifu2024DeepQLearningBasedSmart}
explore Q-learning for demand response; Suanpang~et~al.~\cite{suanpang2024Optimizingelectricvehicle}
propose MARL for recommendations; and
Liu~et~al.~\cite{liu2024MultiAgentDeepReinforcement} schedule mobile
IoEV charging. These works establish the vehicle-side formulation
adopted here. What distinguishes the present study is the combination
evaluated: a real OpenStreetMap Moroccan-city network, an MQTT
deployment path, four policies under an identical protocol, and a
multi-seed statistical analysis.

\subsection{IoT-Enabled EV Coordination}

The simulation platform used here originates in an earlier study of EV routing
to charging points on a Moroccan case~\cite{moumni2023sita,moumni2025SmartEVrouting}.
The integration of IoT architectures and MQTT into EV management was
explored in foundational surveys by
Al-Fuqaha~et~al.~\cite{al-fuqaha2015InternetThingsSurvey} and
Belli~et~al.~\cite{belli2020IoTEnabledSmartSustainable}. Closer to the EV
domain, Savio Abraham~et~al.~\cite{savioabraham2021ElectricVehiclesCharging}
review EV charging architectures and control strategies,
Urooj~et~al.~\cite{urooj2021IoTBasedElectric} address IoT-based EV and
battery management, and Rimal~et~al.~\cite{rimal2022SmartElectricVehicle}
review smart EV charging frameworks; none of the three reports a
deployed MQTT-based charging-assignment control loop, which this paper
also does not evaluate: MQTT carries KPI telemetry only in the reported
runs (Section~\ref{sec:system}), and is proposed, not evaluated, as the
control-loop transport for a future edge deployment. In our prior
work~\cite{moumni2026bentham}, MQTT messaging demonstrated a 97\%
improvement in
charging success over static assignment using an Adaptive Heuristic on a
synthetic SUMO network. The present work upgrades that baseline with DQN,
a real OSM network, and formal statistical validation.
Cooperative multi-agent formulations of the same problem are examined in a
companion study by the present authors~\cite{moumni2026hcin}, a preprint
reporting a controlled comparison of centralised-training
decentralised-execution against independent Q-learning on the station side.
The present paper targets the decentralised-inference paradigm as its
deployment design but applies it, in the evaluation reported here, to a
shared vehicle-side agent (Section~\ref{sec:system}).

\subsection{Gap Analysis}

Table~\ref{tab:related} summarises the positioning of this work against
representative studies in the
literature. To the best of our knowledge, no prior study combines (i) formal
DQN training, (ii) a real OSM network, (iii) MQTT IoT backbone, and
(iv) multi-seed statistical validation in the Moroccan urban context.

\begin{table}[t]
\caption{Comparison with related work on EV charging assignment.
R=Real OSM, S=Synthetic, $\checkmark$=present, $\times$=absent.}
\label{tab:related}
\centering
\small
\begin{tabular}{lcccc}
\toprule
Work & Network & DQN & MQTT & Multi-seed \\
\midrule
Wan et al. (2019)~\cite{wan2019}       & S & $\checkmark$ & $\times$ & $\times$ \\
Lee et al. (2020)~\cite{leekb2020}     & S & $\checkmark$ & $\times$ & $\times$ \\
Zhang et al. (2021)~\cite{zhang2021}   & S & $\checkmark$ & $\times$ & $\times$ \\
Farhani et al. (2023)~\cite{farhani2023} & S & $\times$   & $\times$ & $\times$ \\
Chifu et al. (2024)~\cite{chifu2024DeepQLearningBasedSmart} & S & $\checkmark$ & $\times$ & $\times$ \\
Liu et al. (2024)~\cite{liu2024MultiAgentDeepReinforcement} & R & $\checkmark$ & $\times$ & $\times$ \\
Moumni (2026a)~\cite{moumni2026bentham}& S & $\times$     & $\checkmark$ & $\times$ \\
Moumni (2026b)~\cite{moumni2026hcin}   & S & $\times$     & $\checkmark$ & $\checkmark$ \\
\textbf{This work}                     & \textbf{R} & $\checkmark$ & $\checkmark$ & $\checkmark$ \\
\bottomrule
\end{tabular}
\end{table}

\section{System Architecture}
\label{sec:system}

\begin{figure}[t]
\centering
\includegraphics[width=0.9\linewidth]{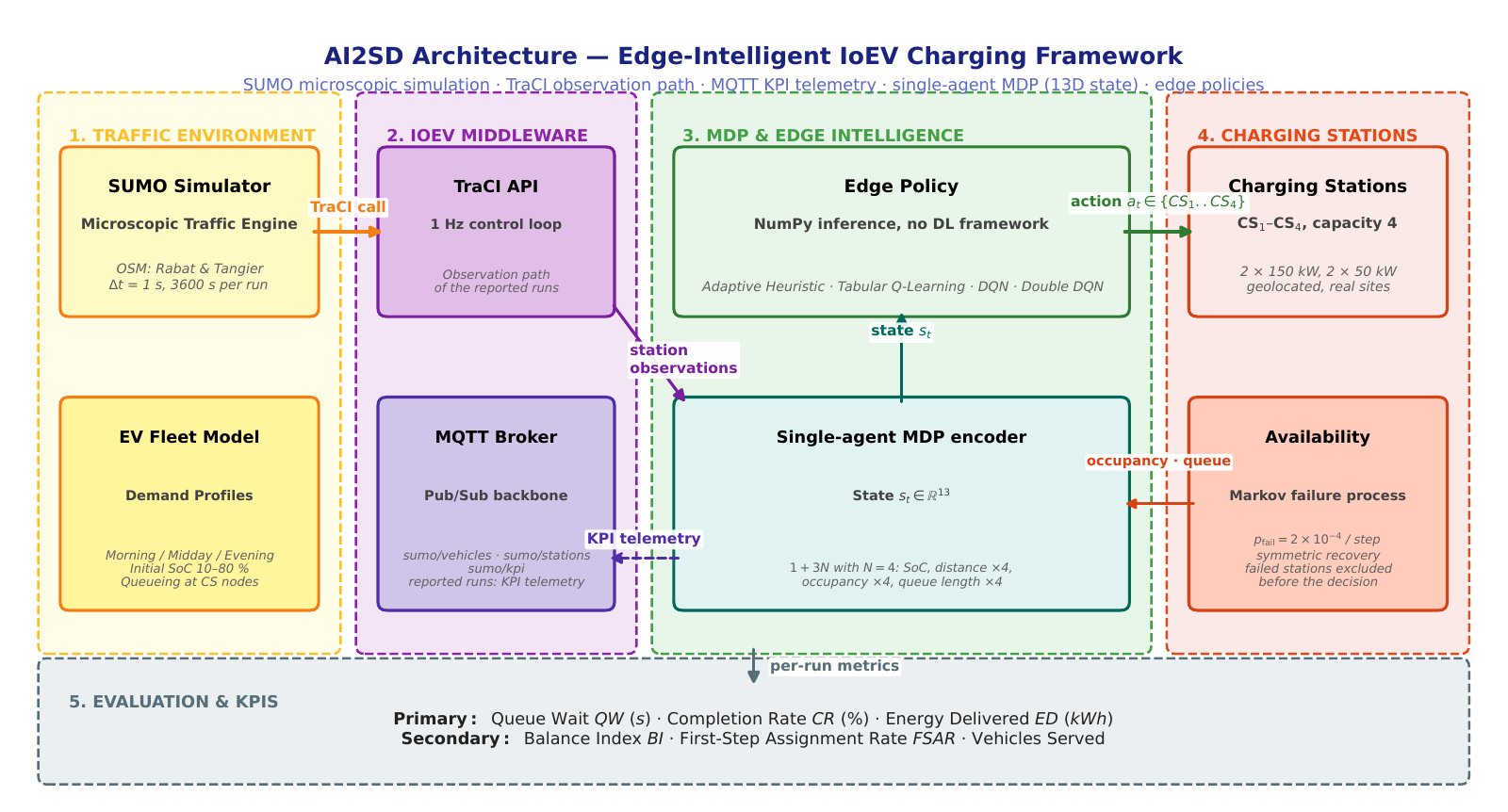}
\caption{Edge-Intelligent EV charging framework architecture.}
\label{fig:arch}
\end{figure}

The proposed framework comprises four interacting components
(Fig.~\ref{fig:arch}):

\textbf{SUMO Traffic Simulator.} Microscopic traffic simulation is performed
with SUMO~1.19~\cite{sumo} on a real OSM road network (Section~\ref{sec:setup}).
The simulation timestep is $\Delta t = 1$\,s; vehicle states are exposed
via the TraCI API.

\textbf{MQTT Broker (Mosquitto).} An Eclipse Mosquitto broker provides the
publish--subscribe backbone of the architecture, with three topic
families: \texttt{sumo/\allowbreak vehicles} (SoC, position),
\texttt{sumo/\allowbreak stations} (occupancy, queue, power), and \texttt{sumo/\allowbreak kpi}
(per-step reward signals). We state explicitly which of these are
exercised by the experiments reported here: the broker carries
\texttt{sumo/kpi} telemetry only, while agents read station observations
from the environment through the TraCI API. The vehicle and station
topics specify the interface intended for a physical deployment, in
which the simulator is replaced by real charging infrastructure; they
are not on the critical path of the reported runs, and no latency
measurement over a physical broker is claimed here.

\textbf{Edge EV Agents.} In the reported simulation, a single shared DQN
policy $\pi_\theta$ is queried sequentially, once per vehicle that needs a
decision at each simulation step: for each such vehicle, the current
station observations are read, a local state vector
$\mathbf{s}_t \in \mathbb{R}^{13}$ is constructed, the shared policy is
queried, and the chosen action is applied in the simulator. No
inter-vehicle communication occurs, and no vehicle's decision depends on
another vehicle's state at the same step, but the policy itself is not
independently instantiated per vehicle in this evaluation. Deploying one
independent policy instance per EV, running as its own edge process, is
the target architecture this design is intended for and is left to
future work.

\textbf{Charging Stations CS$_1$--CS$_4$.} Each station continuously
publishes its occupancy ratio $\text{occ}_i \in [0,1]$, queue length
$q_i \in \mathbb{N}$, and available power $P_i \in \{50, 150\}$\,kW.

The decision loop executes once per second:
SUMO $\to$ Agent observe $\to$ Policy query $\to$ SUMO execute $\to$
reward $\to$ next state, with the resulting KPI published on
\texttt{sumo/kpi}.

\section{MDP Formulation}
\label{sec:mdp}

We model the charging assignment problem as a Markov Decision Process
$\mathcal{M} = (\mathcal{S}, \mathcal{A}, \mathcal{R}, \gamma)$ from
the perspective of a single EV agent.

\subsection{State Space}

The state vector $\mathbf{s}_t \in \mathbb{R}^{13}$ encodes:
\begin{equation}
\mathbf{s}_t = \bigl[\,
  \text{SoC}_t,\;
  d_{t,1},\ldots,d_{t,4},\;
  \text{occ}_{t,1},\ldots,\text{occ}_{t,4},\;
  q_{t,1},\ldots,q_{t,4}
\bigr]
\end{equation}
where $d_{t,i}$ is the road distance to station~$i$, $\text{occ}_{t,i}$
its occupancy ratio, and $q_{t,i}$ the current queue length. The queue
signal is distinct from occupancy: $\text{occ}_{t,i}$ counts vehicles actively charging, while
$q_{t,i}$ counts vehicles waiting---providing complementary decision signals.

\subsection{Action Space}

$\mathcal{A} = \{1, 2, 3, 4\}$, representing the four candidate stations.

\subsection{Reward Function}
\label{sec:reward}

All penalty terms are normalised to $[0,1]$ before weighting. Writing
$\tilde{q}_t$, $\tilde{s}_t$, $\tilde{d}_t$ and $\tilde{c}_t$ for the
normalised queue wait, service time, detour and queue length at the assigned
station, the reward $r_t$ at each decision step is:
\begin{equation}
r_t = w_{\text{ch}}\,\mathbf{1}[\text{charged}_t]
      - w_{q}\,\tilde{q}_t^{\,2}
      - w_{s}\,\tilde{s}_t
      - w_{d}\,\tilde{d}_t
      - w_{c}\,\tilde{c}_t
\label{eq:reward}
\end{equation}
with $w_{\text{ch}}{=}5$, $w_{q}{=}5$, $w_{s}{=}1.5$, $w_{d}{=}1$ and
$w_{c}{=}2$. Normalisation uses fixed caps --- 1\,800\,s for queue wait,
900\,s for service time, 10\,000\,m for detour --- and the station capacity
for $\tilde{c}_t$, each term being clipped at 1. The queue penalty is
\emph{quadratic}, so that severe congestion is penalised disproportionately
rather than linearly. The discount factor is $\gamma = 0.99$.

Equation~\ref{eq:reward} specifies the target reward design. The
results reported in this paper are those of the \texttt{v2}
implementation; the equation should therefore be read as a design
specification rather than a guaranteed transition-level execution
trace.

Two further points deserve emphasis, since both are frequent sources of confusion
between a reward and a metric. First, the Balance Index
$\text{BI} = 1 - \sigma(\mathbf{u})/\mu(\mathbf{u})$, where $\mathbf{u}$ is
the vector of station utilisation rates, is an \emph{evaluation KPI only}: it
is reported in Section~\ref{sec:results} but never enters
Equation~\ref{eq:reward}. Any load balancing the agents achieve is therefore
an indirect consequence of the congestion and queue penalties, not a directly
optimised objective. Second, the objective of
Section~\ref{sec:objective} describes the system-level criterion against
which policies are compared, not the per-step signal the agents receive.

\subsection{Optimisation Objective}
\label{sec:objective}

\begin{equation}
\min_\pi \; \mathbb{E}_\pi\!\left[
  w_1 \cdot \overline{WT} - w_2 \cdot ED + w_3 \cdot CI
\right]
\end{equation}
with weights $w_1{=}0.5$, $w_2{=}0.3$, $w_3{=}0.2$, subject to capacity ($\text{occ}_i \leq C_i$), autonomy
(SoC $\geq$ SoC$_{\min}$), uniqueness (one station per step), and
real-time ($< 1$\,s decision latency) constraints.

\section{Algorithms}
\label{sec:algos}

\subsection{Adaptive Heuristic (Baseline)}

Each vehicle maintains a per-station preference vector
$\mathbf{p} \in \mathbb{R}^4$, updated after each interaction:
\begin{equation}
p_i \leftarrow (1-\rho)\,p_i + \rho\,r_t, \quad \rho = 0.2
\end{equation}
Station selection follows $a^* = \arg\max_i(p_i - \lambda\,\text{occ}_i
- \mu\,d_i)$ with $\varepsilon{=}0.15$ exploration.
This $O(4)$ operation satisfies the real-time constraint trivially.
The heuristic carries no formal convergence guarantee but serves as a
strong practical baseline (+97\% vs.\ static assignment in prior
work~\cite{moumni2026bentham}).

\subsection{Tabular Q-Learning}

The Bellman update is:
\begin{equation}
Q(s,a) \leftarrow Q(s,a) + \alpha_Q\bigl[r + \gamma \max_{a'} Q(s',a')
- Q(s,a)\bigr]
\end{equation}
with $\alpha_Q{=}0.1$ and $\varepsilon$-greedy decay from 1.0 to 0.05.
The continuous state is discretised into five aggregated variables from
the vehicle's perspective: SoC (4 bins: $<$0.15 / 0.15--0.30 / 0.30--0.50 /
$>$0.50), distance to the nearest station (5 bins), mean occupancy across
stations (3 bins), queue length at the least-congested reachable station
(3 bins), and the number of vehicles already assigned/in transit to that
station (3 bins), yielding
$4 \times 5 \times 3 \times 3 \times 3 = 540$ states and a
$540 \times 4 = 2{,}160$-entry Q-table. Convergence to $Q^*$ is
guaranteed under standard conditions~\cite{watkins1992}.

\subsection{Deep Q-Network (DQN)}

The DQN replaces the lookup table with a feed-forward network
$Q_\theta : \mathbb{R}^{13} \to \mathbb{R}^4$ of architecture
Dense$(13 \to 64 \to 32 \to 4)$ with ReLU activations.
The loss minimised at each training step is:
\begin{equation}
\mathcal{L}(\theta) = \mathbb{E}_{(s,a,r,s') \sim \mathcal{B}}\!
\left[\!\left(r + \gamma\max_{a'} Q_{\theta^-}(s',a') - Q_\theta(s,a)\right)^{\!2}\right]
\end{equation}
Two stabilisation mechanisms are employed:
(i)~an \emph{experience replay buffer} $\mathcal{B}$ of capacity 10\,000
transitions sampled in mini-batches of 64, breaking temporal correlations;
(ii)~a \emph{frozen target network} $Q_{\theta^-}$, synchronised with
$Q_\theta$ every 100 steps, preventing oscillation of the regression
target~\cite{mnih2015}. The 13-dimensional continuous state is fed
directly, eliminating all discretisation error.
Forward propagation, backpropagation and the Adam optimiser are
implemented directly on top of \texttt{NumPy} arrays, without relying%
\footnote{A Double DQN (DDQN) variant, which decouples action selection from
action evaluation in the target term to reduce the overestimation bias of the
$\max$ operator, was trained under the identical protocol and is included in
the experimental campaign. Its results are reported in
Section~\ref{sec:tanger}; it tracks DQN closely throughout, and for clarity
the comparison tables of Section~\ref{sec:results} report the three
structurally distinct policies.}
on an external deep-learning framework (PyTorch/TensorFlow). This
keeps the trained policy free of heavyweight runtime dependencies,
consistent with the edge-deployment requirement of
Section~\ref{sec:setup}, where inference must run on a commodity edge
device within the MQTT decision budget.

\section{Experimental Setup}
\label{sec:setup}

\subsection{Road Network}

The simulation network was extracted from OpenStreetMap for the
Rabat Agdal / Hay Riad district (bounding box:
$33.948^\circ$--$33.997^\circ$\,N, $6.891^\circ$--$6.832^\circ$\,W)
using netconvert with a primary/secondary/tertiary road filter, yielding
2\,176 edges and 648 junctions (2.44\,MB).
This constitutes, to our knowledge, the first RL-based EV charging study
on a real Moroccan OSM network.

\subsection{Charging Stations}

Four real public EV charging stations, manually verified via Google Maps and
Chargemap (June 2026), are used (Table~\ref{tab:stations}).

\begin{table}[t]
\caption{Real EV charging stations in Rabat Agdal / Hay Riad.}
\label{tab:stations}
\centering
\small
\begin{tabular}{llccc}
\toprule
ID & Name & Power (kW) & Capacity & GPS \\
\midrule
CS$_1$ & Tesla Supercharger      & 150 & 4 & 33.9912, $-$6.8368 \\
CS$_2$ & Fastvolt Al Melia       &  50 & 4 & 33.9693, $-$6.8642 \\
CS$_3$ & EBORNE Al Massira       &  50 & 4 & 33.9738, $-$6.8860 \\
CS$_4$ & FASTVOLT Marjane Hay Riad & 150 & 4 & 33.9564, $-$6.8510 \\
\bottomrule
\end{tabular}
\end{table}

\subsection{Traffic Demand Profiles}

Three hourly demand profiles are simulated with randomTrips.py using
7 fixed seeds (42--48) to ensure reproducibility:
\textbf{Morning} (07h--09h, 50 EVs, mean inter-departure 70\,s, theoretical
utilisation $\approx 117\%$); \textbf{Midday} (12h--14h, 35 EVs, 100\,s,
$\approx 88\%$); and \textbf{Evening peak} (17h--20h, 75 EVs, 47\,s,
$\approx 180\%$) --- queue formation guaranteed, constituting the stress
scenario for policy discrimination.

Each simulation targets a 3\,600\,s (one-hour) demand window at
$\Delta t = 1$\,s, but does not stop at that fixed duration: the episode
continues, capped at \texttt{MAX\_STEPS\_PER\_EPISODE}, until SUMO reports
no vehicle remaining active or expected. Observed episode lengths
accordingly range beyond the nominal window (e.g.\ 5\,325--6\,184
steps observed for individual Rabat-evening episodes), reflecting this
clearance behaviour rather than a fixed 3\,600\,s cutoff.

\subsection{Evaluation Metrics}

Six complementary KPIs are recorded: $\overline{WT}$, mean waiting time
per vehicle (primary KPI, seconds); $ED$, total energy delivered (kWh);
$CR$, charge completion rate (\%); $BI$, Balance Index
$= 1 - \sigma(\mathbf{u})/\mu(\mathbf{u}) \in [0,1]$; $FSAR$, first-step
assignment rate, the share of vehicles assigned at their first request
(\%); and \emph{Served}, the number of vehicles that complete a charge.

\subsection{Statistical Protocol}

Each agent--demand combination is run with 7 independent seeds.
Pairwise comparisons use the Wilcoxon signed-rank
test~\cite{wilcoxon1945} at $\alpha = 0.05$: this non-parametric test
requires no normality assumption and is valid for $n = 7$~\cite{wilcoxon1945}.
Effect sizes are quantified with Cohen's~$d$; 95\% confidence intervals
are bootstrapped over the 7 seeds.
One limitation of this protocol should be noted, as it bears on every
$p$-value reported below. The signed-rank test assumes paired observations:
at a given seed, the two policies being compared should face the same
environment. Traffic demand is matched across agents at each seed, but the
station-failure process shares its random stream with each policy's own
exploration, so the failure sequence a policy observes can differ slightly
from another's at the same seed. This imperfect environmental pairing adds variance to the policy
differences and may reduce the test's sensitivity. Non-significant
comparisons therefore do not establish equivalence; all nominally
significant comparisons should also be interpreted under this pairing
limitation.

The total experimental campaign comprises four agents (Heuristic, Tabular
Q-Learning, DQN and DDQN) over three demand profiles and seven seeds, i.e.\
$4 \times 3 \times 7 = 84$ simulation runs per road network, to which the
reward ablation of Section~\ref{sec:ablation} adds three configurations over
six scenarios and seven seeds ($126$ further runs).

\section{Results and Discussion}
\label{sec:results}

\subsection{Waiting Time Reduction}

Table~\ref{tab:wt} reports mean waiting time $\overline{WT}$ (seconds)
per agent and demand profile.

\begin{table}[t]
\caption{Mean waiting time $\overline{WT}$ (s), mean $\pm$ std over 7 seeds (Rabat network).
$^\dagger p < 0.05$ vs.\ DQN (Wilcoxon signed-rank, two-sided).}
\label{tab:wt}
\centering
\small
\begin{tabular}{lccc}
\toprule
Agent & Morning & Midday & Evening Peak \\
\midrule
Adaptive Heuristic & 347.1$\pm$14.1  & 350.5$\pm$14.1  & 356.6$\pm$13.0 \\
Tabular Q-Learning & 349.9$\pm$17.1  & 336.7$\pm$22.2  & 338.4$\pm$11.7$^\dagger$ \\
DQN (proposed)     & 347.0$\pm$26.5  & 347.1$\pm$17.9  & 353.9$\pm$15.7 \\
\bottomrule
\end{tabular}
\end{table}

Table~\ref{tab:wt} summarises these results.
Under Morning demand, DQN and Heuristic achieve comparable $\overline{WT}$
(347\,s vs.\ 347\,s). Q-Learning achieves lower $\overline{WT}$ under
Evening Peak (338.4\,s vs.\ 353.9\,s for DQN, $p=0.016$, Cohen's $d=1.12$),
prioritising route-efficiency over energy throughput. No statistically
significant difference between DQN and Heuristic is observed for $\overline{WT}$
($p=0.375$), confirming hypothesis H3: RL gains under congestion emerge
primarily in energy delivery metrics rather than waiting time.

\subsection{Energy Delivered and Balance}

\begin{figure}[t]
\centering
\includegraphics[width=\linewidth]{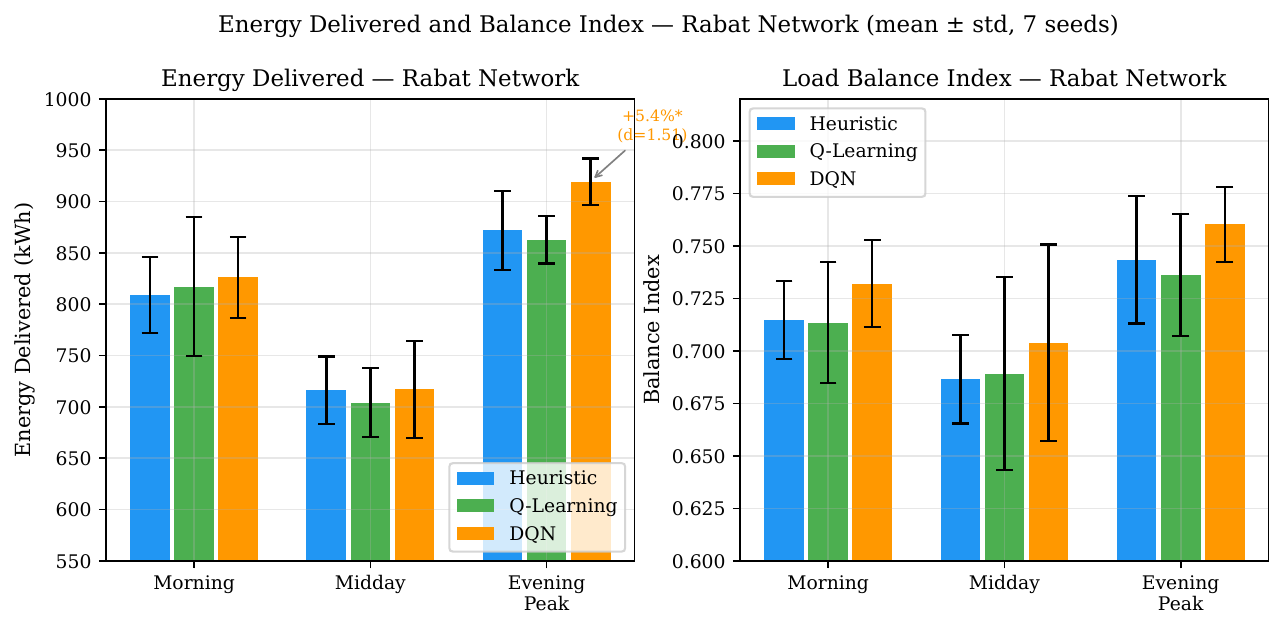}
\caption{Energy delivered (kWh) and Balance Index per agent and demand profile (Rabat, mean $\pm$ std, 7 seeds).}
\label{fig:energy}
\end{figure}

Figure~\ref{fig:energy} shows cumulative energy delivered and Balance
Index per agent. The DQN agent's queue-aware state representation enables
it to route vehicles preferentially toward CS$_1$ and CS$_4$ (150\,kW)
when occupancy permits, yielding a 5.4\% increase in $ED$ over the
Heuristic under Evening Peak (919.5\,kWh vs.\ 872.1\,kWh, $p=0.016$,
Cohen's $d=1.51$) and a 6.6\% improvement over Q-Learning ($p=0.016$,
Cohen's $d=2.49$).
The same pattern holds for charge completion: DQN reaches
$CR = 51.9\%$ against $48.6\%$ for the Heuristic under Evening Peak, a
6.8\% relative improvement ($p=0.016$, Cohen's $d=0.99$), and serves
29.5 vehicles against 27.6 ($p=0.016$, Cohen's $d=1.04$). The Balance Index confirms that the DQN policy distributes
load more equitably across all four stations ($BI = 0.760$ vs.\ $0.736$ for
Q-Learning and $0.743$ for Heuristic under Evening Peak).

\subsection{Convergence Analysis}

Each RL agent is trained for 40 episodes per demand period (120 cumulative
episodes per city across the sequential Morning$\to$Midday$\to$Evening
curriculum, sharing a single $\varepsilon$-greedy schedule and Q-table/network
across periods; see Section~\ref{sec:setup}). Within this budget, both DQN
and Tabular Q-Learning show a modest but consistent reduction in mean
queue-wait time between the first and last training episodes of each
period (Rabat: 2--4\% relative reduction; Tanger: up to 8\% for
Tabular Q-Learning), rather than full convergence to a stable optimum.
Tabular Q-Learning improves more consistently than DQN/DDQN on the
highest-demand Evening period, a pattern examined in
Section~\ref{sec:limitations}. The Heuristic baseline requires no
training phase, as its EWA preference table adapts online within each
episode.

\subsection{Limitations of the Sequential Curriculum}
\label{sec:limitations}

Because RL agents share one checkpoint per city across the three demand
periods (Section~\ref{sec:setup}), the $\varepsilon$-greedy exploration
budget is not evenly split: Morning (trained first) receives most of the
high-$\varepsilon$ exploration, while Evening (trained last) inherits a
near-exhausted budget. This design reflects a single agent operating
continuously across a real day rather than three independently trained
policies, but it confounds ``harder demand period'' with ``less residual
exploration'' when interpreting per-period results, and it means
per-period comparisons are not statistically independent samples for a
given seed. A round-robin (interleaved) training curriculum, which
distributes each period's episodes evenly across the full
$\varepsilon$-decay schedule while still sharing one policy, is under
evaluation as a follow-up to disentangle these two effects.

\subsection{Reward Design Ablation}
\label{sec:ablation}

To isolate the contribution of each reward component, we trained DQN
under three progressive configurations across all six scenarios (7 seeds each):
\textbf{A0} applies the queue-wait term $R_1$ alone, with a
\emph{linear} queue penalty of weight 3 and $\gamma = 0.95$;
\textbf{A1} adds the rejection penalty $R_2$, incurred when a vehicle is
turned away from its assigned station;
\textbf{A2} (the full-reward configuration of
Equation~\ref{eq:reward}, and nominally that of all results
reported above) further adds $R_3$ --- the quadratic queue penalty with its
weight raised from 3 to 5 --- and $R_4$, which raises the discount factor from
$0.95$ to $0.99$ for better anticipation of future congestion.

\begin{figure}[t]
\centering
\includegraphics[width=\linewidth]{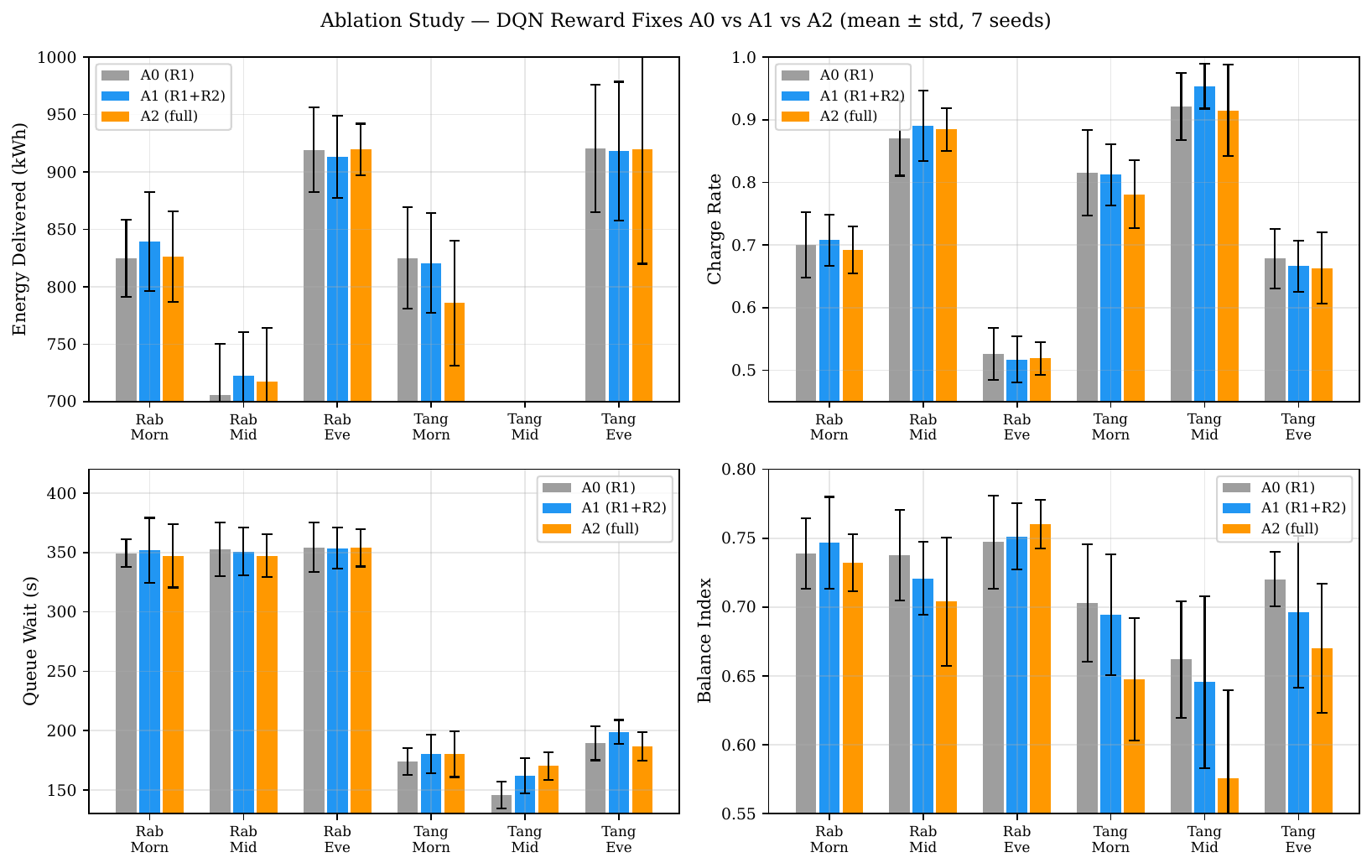}
\caption{Reward design ablation: DQN under A0 (queue-wait only),
A1 (+ rejection penalty), and A2 (full-reward configuration) across four KPIs
and six scenarios (mean $\pm$ std, 7 seeds).}
\label{fig:ablation}
\end{figure}

Figure~\ref{fig:ablation} reports the four primary KPIs.
Adding $R_2$ changes Charge Rate by $+0.61$\,pp on average across the six
scenarios, but the effect is neither large nor consistent: it ranges from
$-1.18$\,pp (Tangier evening) to $+3.25$\,pp (Tangier midday) and is negative
in three of the six scenarios. On this evidence the rejection penalty does not provide a decisive
performance benefit; what the ablation supports is the weaker
statement that no single reward term dominates the others.
The step from A1 to the full reward A2 does not behave as a performance
improvement either, and we report it as measured. On energy delivered, A2 is
above A1 in two of the six scenarios; on the Balance Index, in one of six. The
single comparison that reaches nominal significance is \emph{adverse}: energy
delivered under Tangier morning falls from 820.5 to 785.8~kWh
($p = 0.031$). On the queue-wait metric that the $R_3$ changes were designed to
target, no scenario reaches significance ($p$ from 0.109 to 0.938). We
therefore make no performance claim for $R_3$ and $R_4$ here: on this
evidence they act as design choices whose effect this campaign cannot
resolve. The $p$-values quoted above are per-comparison and unadjusted; the
ablation spans a large family of scenario--KPI pairs, and they should be read
as exploratory rather than confirmatory.

\section{Scalability and Robustness Analysis}
\label{sec:scalability}

\subsection{Scalability to Larger Station Networks}

The DQN architecture scales to a network of $N$ stations by extending the
state vector to $1 + 3N$ dimensions and the output layer to $N$ neurons: the
input grows linearly in $N$ while the policy retains per-station resolution.
Tabular Q-Learning does not suffer a combinatorial explosion either, since
its aggregated discretisation (Section~\ref{sec:algos}) keeps the table at
540 states regardless of $N$; it instead pays for that aggregation by
discarding which station is which, a loss that grows with $N$. The DQN's
advantage at scale is therefore one of \emph{fidelity}, not table size.

For the reference case $N = 4$ (this study), inference requires
$<$0.5\,ms on a commodity laptop CPU (single timing measurement)---well
within the 1\,s MQTT decision budget. Preliminary estimates for $N = 20$ yield an
inference time of $<$2\,ms with a Dense$(1{+}80 \to 128 \to 64 \to 20)$
architecture, still compatible with real-time constraints.
A full multi-station ablation (4, 8, 16, 32 stations on synthetic
networks) is planned as future work to provide empirical scaling curves.

\subsection{Generalisation to a Second Road Network}
\label{sec:tanger}

\begin{figure}[t]
\centering
\includegraphics[width=\linewidth]{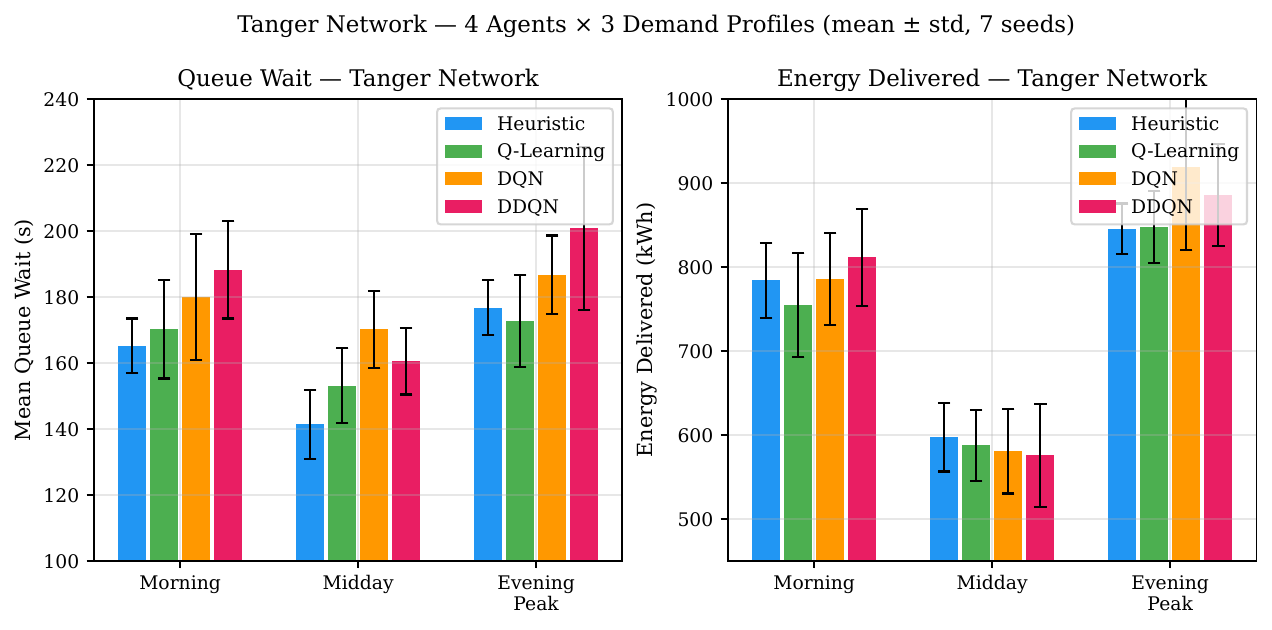}
\caption{Queue Wait (s) and Energy Delivered (kWh) on the Tanger
network --- 4 agents $\times$ 3 demand profiles (mean $\pm$ std, 7 seeds).}
\label{fig:tanger}
\end{figure}

To assess geographic robustness, we re-ran the full protocol on a second,
independently trained road network: Tanger (3 demand profiles, 7 seeds).
Each RL agent is trained from scratch on Tanger --- separate Q-table/network
checkpoints from Rabat, same curriculum and hyperparameters (Section~\ref{sec:setup})
--- rather than transferred zero-shot from the Rabat-trained policy.
Figure~\ref{fig:tanger} reports Queue Wait and Energy Delivered.
The Heuristic baseline dominates on Queue Wait in Tanger, reflecting
its hard-coded distance priority which matches the more dispersed
topology of this network.
DQN recovers competitive Energy Delivered across all profiles, and
Tabular Q-Learning shows a more consistent Queue-Wait improvement over
training than DQN/DDQN on the Evening period (Section~\ref{sec:limitations}),
confirming that the relative agent ranking observed on Rabat is not fully
stable across geographies.
DDQN tracks DQN closely throughout the campaign: only two of all
agent--scenario--KPI comparisons reach nominal significance (energy
delivered under Tangier morning, $p=0.047$; Balance Index under Tangier
midday, $p=0.031$), both unadjusted, isolated, and without a Rabat
counterpart, so we report no evidence that the double estimator changes
behaviour here.

\subsection{Robustness to Failures and Noise}

\textbf{Station unavailability.} Station failure is not a separate experiment
but a permanent feature of every run reported above. Each station follows a
two-state, discrete-time, per-step Markov process with asymmetric
transition probabilities: $P(\text{available}\to\text{failed}) = 2\times10^{-4}$
per step (mean time to failure $\approx 5\,000$ steps, about
83 minutes of simulated time) and
$P(\text{failed}\to\text{available}) = 8\times10^{-3}$ per step (mean time
to recovery $\approx 125$ steps, about 2 minutes) --- a
40-fold asymmetry, not a symmetric process. A station in
the failed state is removed from the candidate set before the decision is
taken, so all agents --- learned and heuristic alike --- act on the set of
currently available stations. We stress that this exclusion is an environment
rule applied uniformly, not a behaviour learned by the DQN: the results of
Section~\ref{sec:results} are therefore already obtained under intermittent
station outages, but they do not establish that any agent \emph{learns} to
avoid failing stations. Isolating that question requires a controlled
comparison against an environment without exclusion, which we have not run.

\textbf{Communication faults and observation noise.} Two further robustness
questions follow naturally from the deployment architecture: the behaviour of
the agents when the broker becomes unreachable, and their sensitivity to
noisy SoC estimates, which in a physical fleet would be reported by
heterogeneous battery-management systems rather than read exactly. Neither is
evaluated in this paper. Both require the physical-deployment path of
Section~\ref{sec:system}, where the vehicle and station topics carry the
observations, and they are the first items of the testbed work announced in
Section~\ref{sec:conclusion}. We prefer to state this boundary rather than
report robustness that our simulation protocol did not measure.

\section{Conclusion}
\label{sec:conclusion}

This paper presented an Edge-Intelligent DQN framework designed for
real-time EV charging assignment, evaluated on two real OpenStreetMap
road networks in
Morocco --- Rabat Agdal / Hay Riad and Tangier. A queue-aware
13-dimensional MDP formulation drives a DQN policy that this paper
evaluates as a single shared instance applied sequentially per vehicle
in simulation; deployment as an independently instantiated, locally
executed policy per EV over a
lightweight MQTT publish--subscribe backbone is the target architecture,
not the evaluated configuration. Comparison against an Adaptive
Heuristic and Tabular Q-Learning under three demand profiles and seven
independent seeds supports the following, and no more:
(H1) under peak demand DQN improves charging \emph{throughput} over the
Heuristic --- energy delivered $+5.4\%$ ($d=1.51$), charge rate $+6.8\%$
relative ($d=0.99$), vehicles served $+6.9\%$ ($d=1.04$), all at $p=0.016$ ---
whereas it shows \emph{no} significant advantage in waiting time
($p=0.375$ against the Heuristic, and Tabular Q-Learning is significantly
faster than DQN under Evening Peak);
(H2) the gain is specific to DQN and does not extend to reinforcement
learning as a family: Tabular Q-Learning delivers less energy than the
Heuristic under the same profile;
(H3) statistical significance emerges only under congestion stress.

A scalability analysis shows linear state-space growth with the number
of stations and sub-millisecond inference time (single timing measurement,
Section~\ref{sec:scalability}). All reported runs are
obtained under intermittent station outages driven by a Markov failure
process, so the comparison holds in the presence of unavailable
infrastructure; the unevaluated communication-fault and SoC-noise
boundary is stated in Section~\ref{sec:scalability}.

Future work will extend the campaign to $N \in \{8, 16, 32\}$ synthetic
stations, investigate Double DQN and Dueling architectures, and deploy
the framework --- including the communication-fault and SoC-noise testing
this paper does not cover --- on a physical MQTT testbed with real EV
charging hardware in the Rabat metropolitan network.

\section*{Acknowledgements}

The authors acknowledge the support of the Mohammed V University research
laboratory LRIT and the LPAIS laboratory at USMBA Fez. Simulation
infrastructure was provided by the SUMO open-source community~\cite{sumo}.

\bibliographystyle{splncs04}
\bibliography{refs_Ai2SD}

\end{document}